\documentclass[10pt]{article} 
\usepackage[preprint]{tmlr}

\usepackage{amsmath,amsfonts,bm}

\def\eqref#1{equation~\ref{#1}}

\def\1{\bm{1}}

\DeclareMathAlphabet{\mathsfit}{\encodingdefault}{\sfdefault}{m}{sl}
\SetMathAlphabet{\mathsfit}{bold}{\encodingdefault}{\sfdefault}{bx}{n}

\usepackage{hyperref}
\usepackage{url}
\usepackage{amssymb}
\usepackage{multirow}
\usepackage{booktabs}   
\usepackage{graphicx}   

\title{Reproducibility Study of "AlphaEdit: Null-Space Constrained Knowledge Editing for Language Models"}

\author{\name Ananth K Suresh\thanks{Equal contribution.} \email ksananth4424@gmail.com \\
\addr Independent 
\AND
\name Arya Hariharan\footnotemark[1] \email arya.hariharan2004@gmail.com \\
\addr Independent
} 
      
\def\month{MM}  
\def\year{YYYY} 
\def\openreview{\url{https://openreview.net/forum?id=XXXX}} 

\begin{document}

\maketitle

\begin{abstract}
\citet{fang2025alphaeditnullspaceconstrainedknowledge} introduced a null-space constrained projection, named AlphaEdit, for locate-then-edit knowledge editing methods, theoretically guaranteeing that edits do not disrupt previously preserved knowledge, and reports substantial gains over existing editing methods on LLaMA3, GPT2-XL, and GPT-J. In this work, we present a reproducibility study of AlphaEdit, reproducing its reported results under the original experimental setup and extending the evaluation along three axes: new model architectures, additional downstream benchmarks, and substantially longer sequential editing horizons. We successfully reproduce AlphaEdit's reported metrics across the original models, though we identify a discrepancy in the reported fluency and consistency metric. Extending AlphaEdit to newer model families, we find that its advantage does not generalize uniformly, which we trace to architectural assumptions in the locate-then-edit paradigm that are violated by these newer models. We further stress-test AlphaEdit's central sequential-editing claim by extending the number of edits well beyond those evaluated in the original paper, and find that performance, which is stable at the originally reported scale, degrades as edits reach a much higher count, indicating that the null-space projection's protection against catastrophic forgetting is bounded rather than unconditional. Finally, we extend evaluation of edited models on three extra benchmarks, namely, BoolQ, HellaSwag, and XSTest and we find that large-scale sequential editing degrades both general downstream task competence and safety-relevant refusal behavior. Our results confirm that AlphaEdit performs as reported within its original scope, while showing that its core theoretical guarantees are sensitive to model architecture and editing scale in ways that have practical implications for its deployment.
\end{abstract}

\section{Introduction}
\label{sec:introduction}
Large language models are known to encode factual errors and outdated knowledge acquired during pretraining, and retraining a model from scratch to correct even a single fact is prohibitively expensive. This has motivated a body of work on model editing which are methods that directly modify a small set of model parameters to update specific facts while leaving the rest of the model's behavior intact. Among these, the locate-then-edit paradigm, which first identifies the parameters most responsible for storing a target fact and then applies a targeted perturbation to them, has become the dominant approach, with methods such as ROME \citep{3600270.3601532} and MEMIT \citep{meng2023memit} demonstrating that individual facts can be edited with a single rank-one or low-rank parameter update.

A persistent weakness of locate-then-edit methods is that the perturbations introduced to encode a new fact are not constrained to leave other, previously correct knowledge untouched. As more edits are applied sequentially, these unconstrained perturbations accumulate and progressively degrade the model's general capabilities. AlphaEdit \citep{fang2025alphaeditnullspaceconstrainedknowledge} claims to address this directly. It projects each parameter perturbation onto the null space of the preserved knowledge before applying it, which the authors prove theoretically guarantees that the model's behavior on preserved facts is unchanged by the edit. Reported on LLaMA3, GPT2-XL, and GPT-J, AlphaEdit is seen to improve the performance of existing locate-then-edit methods.

Despite its strong reported performance, three questions remain. First, like ROME and MEMIT, its implementation relies on model-specific assumptions about parameter naming, layer structure, and subject-token localization, leaving its applicability to newer architectures largely untested. Second, although AlphaEdit is designed for sequential editing, the original evaluation only extends to 3,000 edits, making it unclear whether its null-space protection remains effective at larger scales or eventually reaches a structural limit. Third, the paper focuses on editing-specific metrics such as efficacy, generalization, locality, fluency, and consistency, without evaluating the impact of large-scale editing on broader downstream capabilities or safety-relevant behaviors. In this paper, we present a reproducibility study of AlphaEdit that addresses these three questions. We first reproduce AlphaEdit's reported results under its original experimental setup, to establish a verified baseline before extending it. We then extend the original evaluation along three axes: (1) we apply AlphaEdit to three newer model families, namely Qwen2.5, Gemma-2 and Phi-3, that were not evaluated in the original paper; (2) we extend the number of sequential edits to test whether the null-space guarantee holds at scale; and (3) we evaluate edited models on BoolQ, HellaSwag, and XSTest, three benchmarks not used in the original paper, to assess whether sequential editing measurably degrades general capability and safety behavior.

Our reproduction confirms AlphaEdit's reported efficacy, generalization, and locality scores on the original models, though we identify a discrepancy in the reported consistency metric that we were unable to resolve using the released codebase, which we report as an open issue rather than a confirmed bug. Extending AlphaEdit to Qwen2.5, Gemma-2, Phi-3 and other models with the same family of models as stated in the paper, we find that its advantage does not generalize uniformly. Stress-testing AlphaEdit's sequential editing claim, we find that performance remains stable up to 3000 edits, consistent with the original paper's findings, but degrades after around 5000 edits. This indicates that the null-space projection's protection against catastrophic forgetting, while real, is bounded rather than unconditional. Finally, evaluating edited models on BoolQ, HellaSwag, and XSTest, we find that large-scale sequential editing measurably degrades both general downstream task competence and safety-relevant refusal behavior on models that start with good capability levels.

Our results confirm that AlphaEdit performs as reported within the scope of its original evaluation, while showing that the practical reliability of its theoretical guarantees is sensitive to both model architecture and editing scale in ways that have direct implications for anyone deploying it outside the exact conditions under which it was originally tested. 

\section{Background}
\label{sec:background}

\subsection{Locate-Then-Edit Model Editing}

Transformer-based language models store factual associations in a distributed but partially localizable form. \citet{3600270.3601532} show, using causal tracing, that the recall of a factual association (for example, that the Eiffel Tower is located in Paris) depends disproportionately on a small number of mid-layer MLP modules. They argue that these modules act as key--value memories and the MLP's first projection computes a \emph{key} from the subject token's hidden representation, while the second projection retrieves a corresponding \emph{value} containing the associated fact. This view motivates \emph{locate-then-edit} methods: first identify which layer and which key--value pair encodes a target fact, then directly overwrite the value while leaving the rest of the model untouched. ROME \citep{3600270.3601532} instantiates this idea for single-fact edits. Given a target edit, ROME computes a rank-one update to a chosen MLP down-projection matrix $W$ such that a new key $k_1$ retrieves the desired value $v_1$:

\begin{equation}
\hat{W} = W + \Delta,
\qquad
\Delta =
\frac{(v_1 - Wk_1),k_1^\top}
{k_1^\top k_1}.
\end{equation}

This update is exact for the edited fact but provides no explicit mechanism for preserving other facts stored in the same weight matrix. Edits are applied one at a time, and ROME does not natively support batched or large-scale sequential editing.

MEMIT \citep{meng2023memit} generalizes this approach to many simultaneous edits by distributing the update across multiple layers and solving a least-squares problem over a batch of keys and values rather than a single rank-one update. For a key set

\[
K_1 = [k_1, \dots, k_n]
\]

and target value set $V_1$, MEMIT computes

\begin{equation}
\Delta =
(V_1 - WK_1),
K_1^\top
\left(
K_1 K_1^\top + \lambda I
\right)^{-1},
\end{equation}

where $\lambda$ is a Tikhonov regularization parameter introduced for numerical stability. This formulation allows MEMIT to insert thousands of facts in a single editing pass and to apply edits sequentially in batches. However, like ROME, MEMIT's objective minimizes only the update error on the new facts. It contains no term that explicitly constrains the update's effect on knowledge the model already encodes correctly. In practice, sequential application of MEMIT causes measurable drift in unrelated model behavior as the number of edits grows, a failure mode later characterized as gradual and eventually catastrophic forgetting \citep{gupta2024modeleditingscaleleads}.

\subsection{AlphaEdit: Null-Space Constrained Editing}

AlphaEdit \citep{fang2025alphaeditnullspaceconstrainedknowledge} addresses the preservation problem directly by constraining where in parameter space an edit is allowed to move, rather than only constraining what the edit achieves. The central idea is to project each candidate update onto the left null space of the preserved knowledge so that the update is orthogonal to the directions in which preserved facts are represented. Consequently, the update cannot disturb those facts regardless of its magnitude.

Let $K_0$ denote the key matrix corresponding to knowledge that should be preserved, let $K_1$ and $V_1$ denote the keys and target values of the facts to be edited, and let $K_p$ denote the keys of facts edited in previous rounds of sequential editing. AlphaEdit first computes a projector $P$ onto the null space of $K_0$ using the singular value decomposition of $K_0 K_0^\top$:

\begin{equation}
K_0 K_0^\top = U \Sigma U^\top,
\qquad
P = U_0 U_0^\top,
\end{equation}

where $U_0$ contains the eigenvectors associated with numerically zero eigenvalues. These vectors span the null space of $K_0 K_0^\top$. Any update of the form $\Delta P$ is orthogonal to every key in $K_0$ because

\begin{equation}
K_0^\top P = 0.
\end{equation}

AlphaEdit then solves for the update restricted to this null space:

\begin{equation}
\min_{\tilde{\Delta}}
;
\left|
(W+\tilde{\Delta}P)K_1 - V_1
\right|_F^2
+
\left|
\tilde{\Delta}P K_p
\right|_F^2
+
\lambda
\left|
\tilde{\Delta}P
\right|_F^2.
\end{equation}

The first term fits the new edits, the second penalizes interference with facts introduced in previous editing rounds, and the third is a ridge regularization term. Because the projection matrix $P$ can be precomputed via SVD and the objective is convex and quadratic in $\tilde{\Delta}$, the optimization admits a closed-form solution:

\begin{equation}
\Delta^\star
=
\tilde{\Delta}P
=
R K_1^\top P
\left(
K_p K_p^\top P
+
K_1 K_1^\top P
+
\lambda I
\right)^{-1},
\qquad
R \triangleq V_1 - W K_1.
\end{equation}

The central theoretical claim of AlphaEdit is that because every applied update lies in the null space of $K_0$, the model's behavior on preserved knowledge is provably unaffected by editing. In contrast, preservation in ROME and MEMIT emerges only as an empirical by-product of optimization rather than as a structural guarantee.

Fang et al.~report that this projection step acts as a drop-in addition to existing locate-then-edit pipelines, requiring no modification to how facts are located and changing only how the resulting update is applied. They show that adding the projection improves editing performance over MEMIT, PRUNE \citep{ma2025perturbationrestrainedsequentialmodelediting}, and RECT \citep{gu2024modeleditingharmsgeneral} across LLaMA3, GPT-2 XL, and GPT-J, including under sequential editing of several thousand facts.

This guarantee is nevertheless conditional on assumptions that are not always made explicit in evaluation. First, it presumes that the locate step correctly identifies the subject's key representation. If the wrong token or module is targeted, the resulting update remains orthogonal to the wrong set of directions, and the preservation guarantee does not transfer to the intended knowledge. Second, the null space of $K_0$ is finite-dimensional and is computed once from a fixed sample of preserved knowledge. Consequently, the guarantee is local to that snapshot, and the behavior of the method as $K_p$ accumulates over many rounds of sequential editing, or when applied to architectures differing from those originally evaluated, is not established by the original theoretical result. These motivate the important empirical questions described in Section~\ref{sec:experiments}.

\section{Scope of Reproducibility}
\label{sec:scope}
 
We measure the effect of the null-space projection step by comparing each model's behavior
before and after sequential editing is applied, using the standard editing-specific metrics defined in the
original evaluation protocol. We say that editing is successful for a given fact if the
model assigns higher probability to the target object than to the original object under the editing
prompt and under paraphrases of it, while leaving the model's predictions
on semantically unrelated neighboring facts unchanged. The claims we investigate in
this work are:
 
\begin{itemize}
    \item \textbf{Claim 1:} A model's behavior on preserved knowledge is unaffected by editing using AlphaEdit's technique.
    \item \textbf{Claim 2:} The above guarantee transfers
    without modification to model architectures beyond the three families evaluated in the original
    paper.
    \item \textbf{Claim 3:} The null-space projection's protection against catastrophic forgetting
    continues to hold as the
    number of sequential edits is increased substantially beyond the original range.
    \item \textbf{Claim 4:} The model continues to output coherent and correct text after editing.
\end{itemize}

\section{Experiments}
\label{sec:experiments}

\subsection{Experimental Setup}
\label{sec:setup}


We use the official AlphaEdit codebase without modification, except for resolving dependency issues. Each experiment sequentially edits 2,000 facts in batches of 100 and reports aggregate metrics across all edits. Pre-edit baselines correspond to the unmodified model's performance on the same evaluation prompts. The codebase for the reproducibility study will be made available upon acceptance.

\subsubsection{Editing Datasets}
\label{sec:datasets}
We evaluate on the two benchmarks used in the original paper. CounterFact \citep{3600270.3601532} is a counterfactual knowledge-editing benchmark that pairs
factually true statements with deliberately false target edits (e.g.\ replacing ``Paris'' with a
different city as the capital of France). It is considered the more difficult of the two benchmarks,
since its locality probes are constructed by substituting the edited subject with a closely related
entity sharing the same predicate, which stress-tests whether an edit remains confined to the
intended fact rather than spilling over into semantically adjacent knowledge. ZsRE \citep{levy-etal-2017-zero} is a question-answering benchmark derived via back-translation, providing each example
with a rephrased paraphrase (used to measure generalization) and an out-of-scope neighboring
question (used to measure locality). We use the same evaluation splits and edit ordering protocol as
the original AlphaEdit codebase.

\begin{table}[t]
\centering
\caption{CounterFact results before and after AlphaEdit edits. Eff., Gen., and Spe.\ denote
efficacy, generalization, and specificity, respectively. $\pm$ values are standard errors of the
mean over 2000 cases (SEM\,$=$\,std\,$/\sqrt{2000}$). * denotes reproduced numbers.}
\label{tab:counterfact_repro}
\begin{tabular}{llccccc}
\toprule
\textbf{Method} & \textbf{Model}
  & \textbf{Eff.$\uparrow$}
  & \textbf{Gen.$\uparrow$}
  & \textbf{Spe.$\uparrow$}
  & \textbf{Flu.$\uparrow$}
  & \textbf{Consis.$\uparrow$} \\
\midrule
Pre-edited
  & Llama3-8B
  & $7.80\pm0.60$   & $10.58\pm0.60$  & $89.48\pm0.42$
  & $568.69\pm0.13$ & $3.55\pm0.03$ \\
AlphaEdit
  & Llama3-8B
  & $98.90\pm0.10$  & $94.22\pm0.19$  & $67.88\pm0.29$
  & $622.49\pm0.16$ & $32.40\pm0.11$ \\
AlphaEdit*
  & Llama3-8B
  & $99.25\pm0.19$  & $93.40\pm0.46$  & $68.96\pm0.64$
  & $569.48\pm0.15$ & $3.44\pm0.03$ \\
\midrule
Pre-edited
  & GPT2-XL
  & $22.10\pm0.93$  & $24.45\pm0.83$  & $78.05\pm0.63$
  & $526.45\pm1.02$ & $2.89\pm0.06$ \\
AlphaEdit
  & GPT2-XL
  & $99.50\pm0.24$  & $93.95\pm0.34$  & $66.39\pm0.31$
  & $597.88\pm0.18$ & $39.38\pm0.15$ \\
AlphaEdit*
  & GPT2-XL
  & $99.50\pm0.16$  & $94.60\pm0.41$  & $66.05\pm0.66$
  & $546.99\pm0.82$ & $2.57\pm0.06$ \\
\midrule
Pre-edited
  & GPT-J
  & $15.70\pm0.81$  & $18.10\pm0.76$  & $83.41\pm0.56$
  & $515.41\pm0.76$ & $14.90\pm0.28$ \\
AlphaEdit
  & GPT-J
  & $99.75\pm0.08$  & $96.38\pm0.23$  & $75.48\pm0.21$
  & $618.50\pm0.17$ & $42.08\pm0.15$ \\
AlphaEdit*
  & GPT-J
  & $99.80\pm0.10$  & $96.05\pm0.35$  & $75.70\pm0.59$
  & $522.45\pm0.71$ & $15.44\pm0.24$ \\
\bottomrule
\end{tabular}
\end{table}

\subsubsection{Models}
\label{sec:models}
We reproduce AlphaEdit on the three model families evaluated in the
original paper, namely Llama3-8B \citep{grattafiori2024llama3herdmodels}, GPT2-XL (1.5B) \citep{radford2019language}, and GPT-J (6B) \citep{gpt-j}. We then extend
the evaluation to five additional models not tested in the original paper, spanning a range of
architectures and parameter scales namely Llama3.2 1B and 3B \citep{llama32modelcard}, Qwen2.5-3B \citep{qwen2025qwen25technicalreport},
Phi3-3.8B \citep{abdin2024phi3technicalreporthighly}, and Gemma2-2B \citep{gemmateam2024gemma2improvingopen}. All models are taken from their public Hugging Face checkpoints
without further fine-tuning prior to editing.

\subsubsection{Metrics}
\label{sec:metrics}
Following the original AlphaEdit evaluation protocol, we report the following metrics across both datasets.

\begin{itemize}
\item \textbf{Efficacy}: The proportion of edits for which the model assigns higher probability to the target object than to the original object when evaluated on the editing prompt.

\item \textbf{Generalization}: The same criterion evaluated on paraphrased or rephrased variants of the editing prompt, measuring whether the edit transfers beyond the exact training formulation.

\item \textbf{Specificity}: The proportion of out-of-scope neighborhood prompts for which the model's prediction remains unchanged after editing, measuring the extent to which edits avoid unintended side effects.

\end{itemize}

For CounterFact, we additionally report two more metrics as in the original paper:

\begin{itemize}

\item \textbf{Consistency}: The reference score computed on generated responses to prompts related to the edited fact, measuring whether the model's outputs remain consistent with the target edit across different contexts.

\item \textbf{Fluency}: The generation entropy of model outputs after editing, used as a proxy for the naturalness and linguistic quality of generated text.

\end{itemize}

\subsection{Reproducing AlphaEdit's Original Model Numbers}
As a first experiment, we reproduce AlphaEdit on the three model families evaluated in the
original paper, using the
hyperparameter configurations and editing pipeline released in the official AlphaEdit codebase,
without modification. We apply AlphaEdit sequentially to 2{,}000 edits per model on both
CounterFact and ZsRE with 100 samples per batch (the same setting used in the paper), matching the scale and protocol used in the original paper. For each model and dataset,
we report results both before editing (the pre-edited baseline) and after AlphaEdit is applied. The purpose of this experiment is to establish a
verified baseline, confirming that our reimplementation and evaluation pipeline faithfully reproduce AlphaEdit's reported behavior under its original experimental conditions, before
extending the evaluation to new architectures, benchmarks, and editing scales in the experiments
that follow.

\begin{table}[t]
\centering
\caption{ZsRE results before and after AlphaEdit edits. Eff., Gen., and Spe.\ denote
efficacy, generalization, and specificity, respectively. $\pm$ values are standard errors of the
mean over 2\,000 cases (SEM\,$=$\,std\,$/\sqrt{2000}$). * denotes reproduced numbers.}
\label{tab:zsre_repro}
\begin{tabular}{llccc}
\toprule
\textbf{Method} & \textbf{Model}
  & \textbf{Eff.\%$\uparrow$}
  & \textbf{Gen.\%$\uparrow$}
  & \textbf{Spe.\%$\uparrow$} \\
\midrule
Pre-edited
  & Llama3-8B
  & $37.02\pm0.68$ & $36.33\pm0.67$ & $31.89\pm0.50$ \\
AlphaEdit
  & Llama3-8B
  & $94.47\pm0.13$ & $91.13\pm0.19$ & $32.55\pm0.22$ \\
AlphaEdit*
  & Llama3-8B
  & $94.45\pm0.31$ & $90.68\pm0.44$ & $32.93\pm0.49$ \\
\midrule
Pre-edited
  & GPT2-XL
  & $23.70\pm0.61$ & $22.82\pm0.60$ & $24.97\pm0.55$ \\
AlphaEdit
  & GPT2-XL
  & $94.81\pm0.30$ & $86.11\pm0.29$ & $25.88\pm0.21$ \\
AlphaEdit*
  & GPT2-XL
  & $93.05\pm0.44$ & $83.99\pm0.65$ & $25.97\pm0.58$ \\
\midrule
Pre-edited
  & GPT-J
  & $27.81\pm0.65$ & $27.13\pm0.64$ & $27.53\pm0.58$ \\
AlphaEdit
  & GPT-J
  & $99.79\pm0.14$ & $96.00\pm0.22$ & $28.29\pm0.25$ \\
AlphaEdit*
  & GPT-J
  & $99.62\pm0.09$ & $95.54\pm0.38$ & $28.57\pm0.59$ \\
\bottomrule
\end{tabular}
\end{table}

\subsection{Extending to Newer Models}
As our first extension, we evaluate AlphaEdit on five model families not tested in the original
paper, namely, Llama3.2-1B, Llama3.2-3B, Qwen2.5-3B, Phi3-3.8B, and
Gemma2-2B. These models were chosen to span a range of parameter scales below and around
those of the original three models, as well as a range of architectural departures from the
Llama-style design AlphaEdit's released codebase was originally built around. Phi3-3.8B uses fused
\texttt{qkv\_proj} and \texttt{gate\_up\_proj} attention and MLP projections in place of separate
projection matrices, while Gemma2-2B uses a GeGLU MLP variant together with a SentencePiece-derived
tokenizer, in contrast to the BPE tokenizers used by all three original models. We apply AlphaEdit
sequentially on both CounterFact and ZsRE, under the same evaluation
protocol and metrics used in our reproduction of the original three models.

\subsection{Increasing the Number of Edits}
\label{sec:edit-increase}
In the original paper, the authors additionally  evaluate the intrinsic knowledge of post-edited LLMs through General Capability Tests using six natural language tasks from the General Language Understanding Evaluation (GLUE) \citep{wang-etal-2018-glue}. which consists of several tasks, namely the Stanford Sentiment Treebank (SST) \citep{socher-etal-2013-recursive}, Microsoft Research Paraphrase Corpus (MRPC) \citep{dolan-brockett-2005-automatically}, Massive Multi-task Language Understanding (MMLU) \citep{hendryckstest2021}, (Recognizing Textual Entailment (RTE) \citep{10.1007/11736790_9}, Corpus of Linguistic Acceptability (CoLA) \citep{warstadt-etal-2019-neural} and Natural Language Inference (NLI) \citep{williams-etal-2018-broad}. We reproduce these results. However, we stress-test AlphaEdit's central sequential-editing claim by evaluating
its performance well beyond the 3000 edit counts considered in the original paper. Rather than evaluating
only at a single fixed scale, we apply AlphaEdit sequentially in increments of 1{,}000 edits, up to a
maximum of 10{,}000 edits, evaluating model performance after every increment (i.e.\ at 1{,}000,
2{,}000, 3{,}000, \ldots, 10{,}000 edits). This finer-grained, multi-point evaluation allows us to
trace the full trajectory of AlphaEdit's performance as the number of accumulated sequential edits
grows, rather than only comparing two or three isolated edit-count snapshots, and makes it possible
to identify the specific point, if any, at which the null-space projection's protection against
catastrophic forgetting or capability begins to break down. We use the same three models as in our
reproduction of AlphaEdit's original results, along with the Llama 3.2 3B model, applying edits cumulatively and sequentially within
each run.

\subsection{Additional Benchmarks}
Finally, we evaluate the edited models on three other benchmarks not used in the original
paper. BoolQ \citep{clark-etal-2019-boolq} is a yes/no question-answering benchmark used to
assess general reading-comprehension capability, HellaSwag \citep{zellers-etal-2019-hellaswag} is a
sentence-completion benchmark used to assess commonsense reasoning and XSTest \citep{rottger-etal-2024-xstest} is a safety benchmark consisting of both genuinely unsafe prompts that a model should
refuse and superficially sensitive-sounding but benign prompts that a model should answer, used to
assess whether a model's refusal behavior remains correctly calibrated. We evaluate BoolQ and
HellaSwag using standard zero-shot accuracy, and XSTest using the binary should-answer/should-refuse behavior associated with each prompt, comparing the model's actual response behavior
against this expected label. We evaluate on these three benchmarks up to 10,000 edits, at 1000 edit intervals, to track how general
capability and safety-relevant refusal behavior change as the number of accumulated edits increases. We use the same four models as in Section~\ref{sec:edit-increase} and the purpose of this
experiment is to test whether degradation in general capability and safety
behavior emerges at a point along the editing trajectory that isn't tracked in the original paper. 

\begin{table}[t]
\centering
\caption{CounterFact results before and after AlphaEdit edits. Eff., Gen., and Spe.\ denote efficacy, generalization, and specificity, respectively. $\pm$ values are standard errors of the mean over 2000 cases (SEM $=$ std $/ \sqrt{2000}$). * represents extended work during reproducibility study.}
\label{tab:counterfact_new_models}
\begin{tabular}{llccccc}
\toprule
\textbf{Method} & \textbf{Model}
& \textbf{Eff.$\uparrow$}
& \textbf{Gen.$\uparrow$}
& \textbf{Spe.$\uparrow$}
& \textbf{Flu.$\uparrow$}
& \textbf{Consis.$\uparrow$} \\
\midrule

Pre-edited
& Llama3.2-1B
& $15.25\pm0.80$
& $17.00\pm0.73$
& $85.12\pm0.52$
& $543.74\pm0.16$
& $2.65\pm0.03$ \\
AlphaEdit*
& Llama3.2-1B
& $91.70\pm0.62$
& $87.00\pm0.63$
& $59.97\pm0.74$
& $530.51\pm0.20$
& $4.80\pm0.05$ \\

\midrule

Pre-edited
& Llama3.2-3B
& $11.20\pm0.71$
& $13.63\pm0.66$
& $87.14\pm0.48$
& $504.49\pm0.58$
& $2.28\pm0.03$ \\
AlphaEdit*
& Llama3.2-3B
& $97.45\pm0.35$
& $90.92\pm0.54$
& $66.46\pm0.68$
& $482.62\pm0.58$
& $1.77\pm0.02$ \\

\midrule

Pre-edited
& Phi3-3.8B
& $46.35\pm1.12$
& $44.98\pm0.81$
& $54.46\pm0.99$
& $334.06\pm1.54$
& $4.38\pm0.08$ \\
AlphaEdit*
& Phi3-3.8B
& $48.85\pm1.12$
& $49.95\pm0.81$
& $49.70\pm1.06$
& $325.43\pm1.52$
& $1.82\pm0.06$ \\

\midrule

Pre-edited
& Qwen2.5-3B
& $15.65\pm0.81$
& $18.77\pm0.75$
& $84.01\pm0.54$
& $328.82\pm2.72$
& $6.37\pm0.18$ \\
AlphaEdit*
& Qwen2.5-3B
& $99.80\pm0.10$
& $88.78\pm0.58$
& $79.97\pm0.57$
& $332.13\pm2.74$
& $6.84\pm0.18$ \\

\midrule

Pre-edited
& Gemma2-2B
& $12.8\pm0.75$
& $16.6\pm0.72$
& $86.08\pm0.50$
& $257.59\pm0.76$
& $0.18\pm0.01$ \\
AlphaEdit*
& Gemma2-2B
& $45.25\pm1.11$
& $48.18\pm0.96$
& $55.24\pm0.82$
& $248.05\pm0.72$
& $0.12\pm0.003$ \\

\bottomrule
\end{tabular}
\end{table}

\section{Results and Discussion}
\label{sec:results}

\subsection{Results of Reproducing Original Model Numbers}
Table~\ref{tab:counterfact_repro} and Table~\ref{tab:zsre_repro} report our reproduction of AlphaEdit.

Our reproduction of AlphaEdit on the CounterFact benchmark recovers the primary findings reported in the original paper. Across all three evaluated models (Llama3-8B, GPT2-XL, and GPT-J), AlphaEdit increases efficacy from below 25\% in the pre-edited models to approximately 99\%, while simultaneously achieving strong generalization performance exceeding 93\%. The observed reduction in specificity is consistent with the original study.While efficacy, generalization, and specificity closely match the reported values across all three models, we observe systematic discrepancies in the fluency and consistency metrics. Fluency is 8–15\% lower than reported, and consistency is lower by a factor of roughly 3–10× depending on the model. Both discrepancies were traced to the fact that the model was collapsing and not generating coherent output for any of the models. We describe more on this in ~\ref{sec:output-quality}. Despite these discrepancies in generation-based metrics, the probability-based metrics, which are the primary evidence for AlphaEdit's core claims, are reproduced well. The results provide strong support for the main conclusions of the paper. Overall, the reproduced results provide strong evidence supporting the main conclusions of the AlphaEdit paper. The fluency gap is consistent with the \texttt{generate\_fast()} caching issue described in Section \ref{sec:output-quality}, entropy-based fluency depends on generation quality, and the batch-padding misalignment suppresses it independently of the probability-based editing metrics.

For the ZsRE benchmark, the reproduced results closely match those reported in the original paper. Across all three models, AlphaEdit consistently achieves large gains in efficacy and generalization while maintaining stable specificity. The substantial increase in efficacy demonstrates that the edited facts are successfully incorporated into the model. The similarly large improvements in generalization indicate that the edited knowledge transfers to paraphrased queries rather than being restricted to the exact editing prompt. Furthermore, the near-constant specificity scores suggest that the editing procedure preserves unrelated neighborhood knowledge despite the accumulation of 2000 sequential edits. This method is highly reproducible and its reported improvements do not seem to be sensitive to minor implementation or experimental variations.

\subsection{Results of Testing on Additional Models}


\begin{table}[t]
\centering
\caption{ZsRE results before and after AlphaEdit edits. Eff., Gen., and Spe. denote edit efficacy, generalization, and specificity, respectively. $\pm$ values are standard errors of the mean over 2000 cases (SEM $=$ std $/ \sqrt{2000}$). * represents extended work during reproducibility study.}
\label{tab:zsre_new_models}
\begin{tabular}{llccc}
\toprule
\textbf{Method} & \textbf{Model}
& \textbf{Eff. $\uparrow$}
& \textbf{Gen. $\uparrow$}
& \textbf{Spe. $\uparrow$} \\
\midrule
Pre-edited
& Llama3.2-1B
& $29.85\pm0.62$
& $29.18\pm0.62$
& $26.40\pm0.48$ \\
AlphaEdit*
& Llama3.2-1B
& $88.04\pm0.49$
& $82.38\pm0.61$
& $24.92\pm0.46$ \\
\midrule
Pre-edited
& Llama3.2-3B
& $32.66\pm0.64$
& $31.64\pm0.63$
& $28.19\pm0.49$ \\
AlphaEdit*
& Llama3.2-3B
& $92.88\pm0.35$
& $87.15\pm0.52$
& $29.62\pm0.49$ \\
\midrule
Pre-edited
& Phi3-3.8B
& $2.15\pm0.15$
& $2.23\pm0.15$
& $3.23\pm0.17$ \\
AlphaEdit*
& Phi3-3.8B
& $0.00\pm0.00$
& $0.00\pm0.00$
& $0.01\pm0.01$ \\
\midrule
Pre-edited
& Qwen2.5-3B
& $40.93\pm0.68$
& $38.73\pm0.67$
& $37.23\pm0.59$ \\
AlphaEdit*
& Qwen2.5-3B
& $96.81\pm0.27$
& $90.02\pm0.50$
& $46.26\pm0.67$ \\
\midrule
Pre-edited
& Gemma2-2B
& $10.85\pm0.37$
& $10.17\pm0.36$
& $12.39\pm0.28$ \\
AlphaEdit*
& Gemma2-2B
& $0.43\pm0.07$
& $0.28\pm0.05$
& $0.50\pm0.08$ \\
\bottomrule
\end{tabular}
\end{table}



Tables~\ref{tab:counterfact_new_models} and \ref{tab:zsre_new_models} report AlphaEdit's performance across five modern architectures, revealing highly model-dependent generalization. Qwen2.5-3B and the Llama3.2 family achieve the strongest results: Qwen2.5-3B's CounterFact efficacy jumps to 99.80\%. Notably, on the ZsRE benchmark, these successful models achieve >92\% rewrite and >87\% paraphrase accuracies after 2,000 sequential edits while fully preserving neighborhood accuracy, cleanly avoiding the specificity degradation typically observed in the CounterFact setting.

In contrast, AlphaEdit performs substantially worse than on the original architectures for the models Phi3-3.8B and Gemma2-2B. Targeted diagnostics of the locate-then-edit pipeline reveal these failures may stem from structural incompatibilities rather than superficial errors. For Phi3-3.8B, evaluation pipelines operate correctly, but the editing procedure yields an algorithmic null result. The model's base probability machinery produces a near-random signal for target completions (e.g., $P(\text{" Paris"}) \approx P(\text{" Berlin"})$), which persists post-editing. Architecturally, Phi3 employs fused \texttt{qkv\_proj} and \texttt{gate\_up\_proj} matrices. This divergence from the Llama-style weight structure may likely break the subject-token localization heuristic, causing the null-space projection to be computed over a geometrically irrelevant subspace.

For Gemma2-2B, we initially produced near-zero metrics on both benchmarks. These were most likely partially caused by an evaluation artifact where the SentencePiece tokenizer prepends a BOS token to targets, misaligning logit indexing. Correcting this ensured the base probability machinery was sound ($P(\text{" Paris"}) = 0.20$ vs. $P(\text{" Berlin"}) = 0.0001$). However, an algorithmic failure still persisted where the corrected CounterFact efficacy reaches only 45.25\%, and ZsRE rewrite accuracy collapses to 0.43\%. We traced this to Gemma2's \texttt{post\_feedforward\_layernorm}. AlphaEdit's closed-form $v^*$ optimization assumes the weight update $\Delta W$ contributes linearly to the residual stream. Gemma2's post-MLP RMSNorm might non-linearly rescale this perturbation, fundamentally invalidating the null-space projection's mathematical assumptions.

Ultimately, AlphaEdit does not seem like a universal drop-in solution. Extending it reliably to diverse architectures may require adapting localization heuristics for fused projections and modifying the core optimization objective to account for intervening layer normalizations.

\begin{table}[t]
\centering
\caption{CounterFact performance before editing and after cumulative AlphaEdit edits.
Eff., Gen., and Spe.\ denote efficacy, generalization, and specificity, respectively.
Reported values are mean $\pm$ standard error of the mean (SEM).}
\label{tab:counterfact_edit_scaling}
\begin{tabular}{llccccc}
\toprule
\textbf{Model} & \textbf{Setting}
& \textbf{Eff.$\uparrow$}
& \textbf{Gen.$\uparrow$}
& \textbf{Spe.$\uparrow$}
& \textbf{Flu.$\uparrow$}
& \textbf{Consis.$\uparrow$} \\
\midrule

\multirow{4}{*}{Llama3.2-3B}
& Pre-edited
& $11.20\pm0.71$
& $13.63\pm0.66$
& $87.14\pm0.48$
& $504.49\pm0.58$
& $2.28\pm0.03$ \\
& 2,000 edits
& $97.45\pm0.35$
& $90.92\pm0.54$
& $66.46\pm0.68$
& $482.62\pm0.58$
& $1.77\pm0.02$ \\
& 5,000 edits
& $48.24\pm0.71$
& $47.50\pm0.53$
& $46.72\pm0.59$
& $320.36\pm0.94$
& $4.49\pm0.05$ \\
& 10,000 edits
& $47.51\pm0.50$
& $47.08\pm0.39$
& $44.86\pm0.42$
& $320.87\pm0.68$
& $4.89\pm0.04$ \\
\midrule

\multirow{4}{*}{Llama3-8B}
& Pre-edited
& $7.80\pm0.60$
& $10.58\pm0.60$
& $89.48\pm0.42$
& $568.69\pm0.13$
& $3.55\pm0.03$ \\
& 2,000 edits
& $99.25\pm0.19$
& $93.40\pm0.46$
& $68.96\pm0.64$
& $569.48\pm0.15$
& $3.44\pm0.03$ \\
& 5,000 edits
& $49.02\pm0.71$
& $49.74\pm0.65$
& $49.49\pm0.64$
& $324.20\pm0.96$
& $3.06\pm0.03$ \\
& 10,000 edits
& $69.35\pm0.46$
& $60.75\pm0.41$
& $52.86\pm0.33$
& $533.54\pm0.11$
& $2.51\pm0.01$ \\
\midrule

\multirow{4}{*}{GPT2-XL}
& Pre-edited
& $22.10\pm0.93$
& $24.45\pm0.83$
& $78.05\pm0.63$
& $526.45\pm1.02$
& $2.89\pm0.06$ \\
& 2,000 edits
& $99.50\pm0.16$
& $94.60\pm0.41$
& $66.05\pm0.66$
& $546.99\pm0.82$
& $2.57\pm0.06$ \\
& 5,000 edits
& $49.22\pm0.71$
& $49.09\pm0.58$
& $51.31\pm0.64$
& $364.28\pm1.13$
& $3.38\pm0.03$ \\
& 10,000 edits
& $48.48\pm0.50$
& $48.67\pm0.41$
& $51.84\pm0.45$
& $360.56\pm0.82$
& $3.09\pm0.02$ \\
\midrule

\multirow{4}{*}{GPT-J}
& Pre-edited
& $15.70\pm0.81$
& $18.10\pm0.76$
& $83.41\pm0.56$
& $515.41\pm0.76$
& $14.90\pm0.28$ \\
& 2,000 edits
& $99.80\pm0.10$
& $96.05\pm0.35$
& $75.70\pm0.59$
& $522.45\pm0.71$
& $15.44\pm0.24$ \\
& 5,000 edits
& $48.28\pm0.71$
& $48.90\pm0.56$
& $51.41\pm0.66$
& $311.50\pm0.99$
& $4.74\pm0.04$ \\
& 10,000 edits
& $97.83\pm0.15$
& $89.66\pm0.24$
& $62.18\pm0.28$
& $533.28\pm0.33$
& $17.25\pm0.12$ \\
\bottomrule
\end{tabular}
\end{table}

\subsection{Results of Extending the Edit Count}
\begin{figure}[t]
    \centering
    \includegraphics[width=0.99\textwidth]{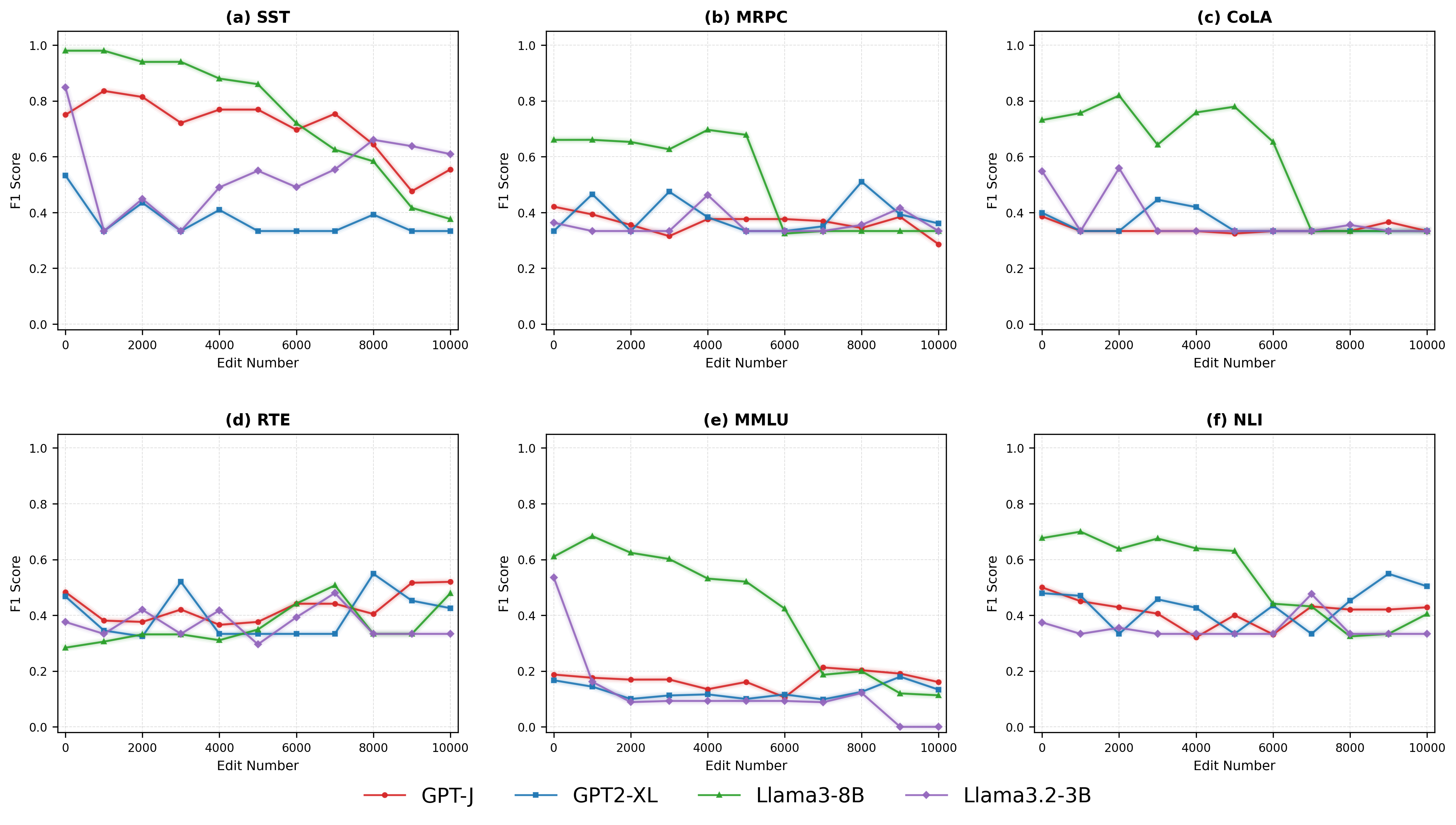}
    \caption{General capability preservation of AlphaEdit under large-scale sequential editing. F1 scores on SST, MRPC, CoLA, RTE, MMLU, and NLI are reported. While AlphaEdit generally avoids catastrophic degradation, performance exhibits model-dependent trends and gradual decline at higher edit counts.}
    \label{fig:extended_edit_count}
\end{figure}

We reproduced the general capability evaluation presented in the original paper and extended the study by increasing the number of sequential edits from the original range of up to 3,000 edits to 10,000 edits. Figure~\ref{fig:extended_edit_count} presents the resulting F1 scores on the six downstream tasks as the number of accumulated edits increases. We also test on the CounterFact dataset at 5000 and 10000 edits, shown in Table~\ref{tab:counterfact_edit_scaling}. In the original work, AlphaEdit demonstrated remarkable stability, with performance remaining largely unchanged even after 3000 edits. Our results reproduce this trend, but also reveal several model-dependent behaviors that become apparent only at larger edit counts.

Table~\ref{tab:counterfact_edit_scaling} shows that degradation onset and recovery on CounterFact are strongly architecture-dependent. All four models maintain high efficacy through 2,000 edits, but diverge thereafter. Llama3.2-3B and GPT2-XL exhibit the most persistent degradation. Llama3-8B follows a similar trend initially but partially recovers by 10,000 edits. GPT-J shows the strongest recovery, rebounding almost fully after a drop at 5,000 edits. This suggests that the effectiveness of null-space projection at high edit counts is not uniformly bounded and that GPT-J's weight structure may align particularly well with AlphaEdit's geometric assumptions. While the shared decline around 5,000 edits points to a common structural limit, potentially effective null-space saturation, the differing recovery patterns indicate that this limit interacts differently with each architecture. Whether GPT-J's recovery reflects genuine re-stabilization or an evaluation artifact remains an open question.

A similar pattern is observed on GLUE. GPT-J demonstrates the greatest robustness to long edit sequences, with gradual degradation and partial recovery at higher edit counts. Some tasks, such as MMLU and CoLA, start from relatively low baselines, however. GPT2-XL remains stable on several tasks, particularly MRPC, RTE, and NLI, although its overall scores are generally lower than GPT-J's. Importantly, neither model experiences catastrophic collapse, even after 10,000 edits. Llama3-8B achieves the strongest initial performance across most tasks, substantially outperforming GPT-J and GPT2-XL at low edit counts. However, several tasks, including SST, CoLA, MMLU, and NLI, show steady degradation beyond roughly 5,000–7,000 edits. The decline is most severe on MMLU, which falls from approximately 0.6 F1 to near 0.1 by 10,000 edits, suggesting that knowledge-intensive capabilities are particularly sensitive to prolonged editing. Nevertheless, unlike competing editing methods reported in the original paper, AlphaEdit maintains non-zero performance and degrades gradually rather than catastrophically.

Compared with the original paper, two observations emerge. First, AlphaEdit's central claim of avoiding the rapid capability collapse seen in methods such as MEMIT, ROME, and pruning-based approaches is largely supported. None of the evaluated models exhibit the abrupt drop to near-zero performance characteristic of prior methods. Second, extending evaluation to 10,000 edits reveals degradation effects that are not visible within the original 3,000-edit range. Although AlphaEdit remains substantially more robust than previous approaches, its ability to preserve general capabilities is not indefinite and varies considerably across model architectures.

\subsection{Results of Evaluating on Additional Benchmarks} 
\begin{figure}[t]
    \centering
    \includegraphics[width=0.99\textwidth]{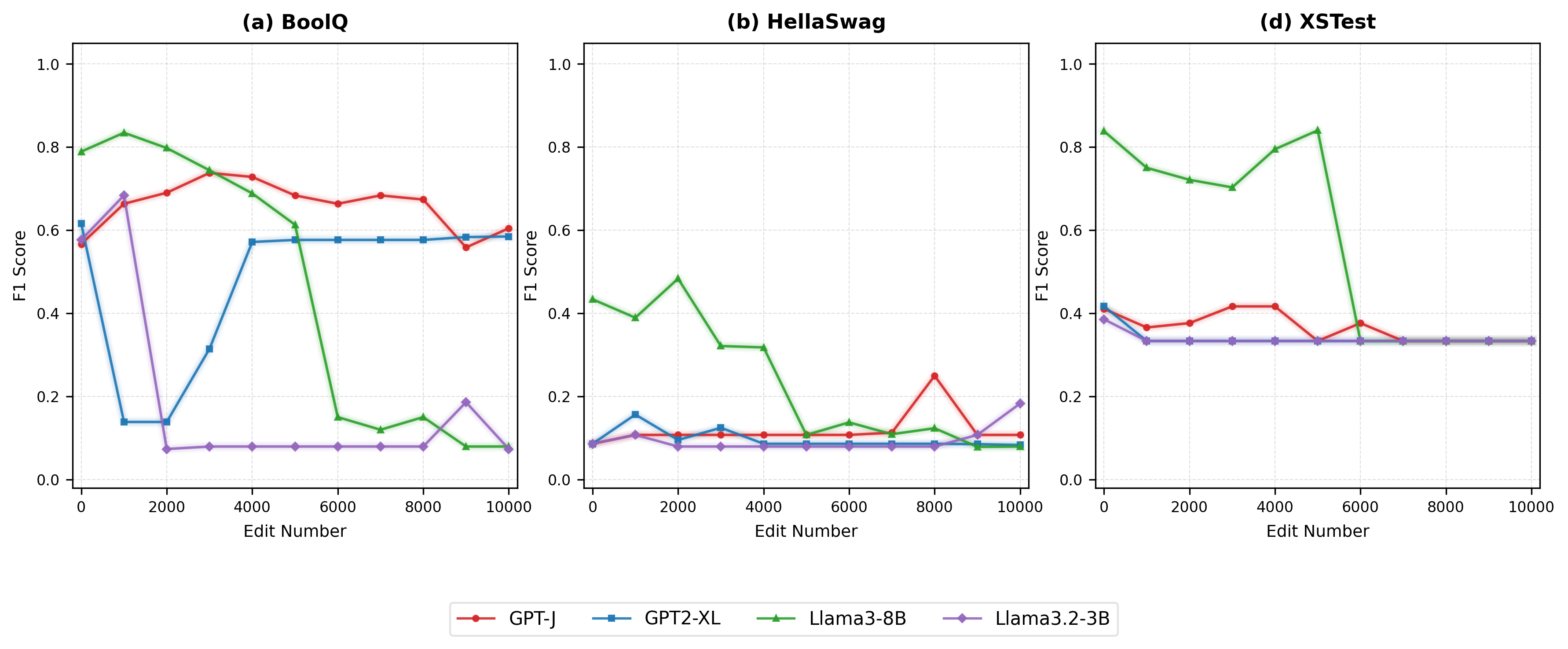}
    \caption{Additional capability benchmark tests. F1 scores are reported as the number of edits increases from 0 to 10,000 for GPT-J, GPT2-XL, Llama3-8B, and Llama3.2-3B. For models that start out with good capability scores, a degradation is seen after higher number of edits.}
    \label{fig:additional-benchmarks}
\end{figure}

Figure~\ref{fig:additional-benchmarks} reports the F1 scores obtained after up to 10,000 sequential edits across GPT-J, GPT2-XL, Llama3-8B, and Llama3.2-3B on the additional benchmarks tested.

\subsubsection{BoolQ}
BoolQ exhibits the greatest variation across models. GPT-J demonstrates the strongest robustness. GPT2-XL experiences a substantial degradation after the first 1,000-2,000 edits but subsequently recovers and stabilizes near its initial performance level. In contrast, both Llama models show significant deterioration as the number of edits increases. Llama3-8B initially achieves the highest performance among all models, exceeding 0.8 F1, but undergoes a sharp decline after approximately 5,000 edits and falls below 0.1 F1 by the end of the experiment. Similarly, Llama3.2-3B collapses rapidly after the first few thousand edits and remains at a low performance level thereafter. 

\subsubsection{HellaSwag}
Performance on HellaSwag remains relatively low for all models, and the benchmark may be challenging for the models under the chosen evaluation setting. GPT-J and GPT2-XL remain poor performing but stable across all edit counts, exhibiting only minor fluctuations. Llama3-8B begins with substantially higher performance than the other models but gradually degrades as edits accumulate, eventually converging toward the performance level of the GPT-based models. Llama3.2-3B remains largely unchanged throughout the experiment, aside from a small increase at the highest edit count. On models like Llama3-8B with good reasoning capabilities, a trend to expect is for capability to degrade after a larger number of edits.

\subsubsection{XSTest}
GPT2-XL and Llama3.2-3B remain almost perfectly stable after an initial drop, maintaining a nearly constant score throughout the remaining edits. GPT-J exhibits minor fluctuations but retains performance within a narrow range. Llama3-8B again starts with the highest performance and remains stable until approximately 5,000 edits, after which a sharp decline occurs and performance converges to the level achieved by the other models. The convergence of all models to a similar score at high edit counts suggests that robustness-oriented capabilities may be less sensitive to editing than knowledge-intensive capabilities, although degradation is still evident in larger models. The degradation patterns differ across benchmark categories. BoolQ, which relies heavily on factual knowledge and reading comprehension, shows the largest performance losses, whereas HellaSwag and XSTest remain comparatively stable for models that generally have low capability to begin with. For larger models with more capability, repeated edits are shown to create definitive drops in capability.

\subsection{Output Quality}
\label{sec:output-quality}
During our reproduction of the AlphaEdit experiments, we identified a potential 
source of discrepancy in the consistency (reference) score computation and the fluency metric. The custom 
\texttt{generate\_fast()} function used for text generation initializes the context 
window to the minimum prompt length across the batch, which may cause misaligned 
KV cache construction for longer prompts, potentially resulting in incoherent 
outputs for models such as GPT-J and Llama3-8B. Replacing \texttt{generate\_fast()} 
with HuggingFace's \texttt{model.generate()} and processing prompts individually 
appeared to resolve this issue for the pre-edited baseline, bringing consistency 
scores closer to the paper's reported values. However, post-edit consistency scores 
remained low and the generated text continued to exhibit incoherence. This might
suggest that the editing process could be degrading the model's text 
generation capability in ways not captured by the probability-based metrics 
(efficacy, generalization, and specificity), all of which matched the paper's 
reported values throughout. This still remains an open reproducibility issue that could not be solved through the repository details alone.

\section{Conclusion}
\label{sec:conclusion}
This work presented a reproducibility study of AlphaEdit, confirming its key claims while exposing  operational boundaries across modern network architectures and expanded operational scales. Our findings validate that AlphaEdit delivers high editing efficacy and strong semantic generalization within its originally proposed configuration of 2,000 edits on LLaMA3-8B, GPT2-XL, and GPT-J models. However, our replication identified a persistent discrepancy in text-generation performance, revealing an underlying structural output collapse that significantly suppresses baseline consistency and fluency metrics. Furthermore, our systematic extensions demonstrate that AlphaEdit's advantage does not generalize uniformly to newer architectural designs. The technique fails on architectures like Phi-3-3.8B and Gemma-2-2B potentially due to structural departures from standard Llama-style layouts, such as fused parameter matrices and intermediate non-linear layer normalization modules that break foundational null-space orthogonality constraints. Finally, stress-testing the framework against an extended multi-point trajectory up to 10,000 edits reveals that AlphaEdit's protection against catastrophic forgetting is fundamentally bounded rather than unconditional. Accumulating edits past a threshold of 5,000 updates triggers a steep performance degradation while progressively eroding the model's general downstream capability and calibrated safety refusal behavior.

\section{Limitations}
\label{sec:limitations}
While this study comprehensively evaluates AlphaEdit, certain scoping boundaries present opportunities for future research. First, our diagnostics isolate structural incompatibilities via mathematical constraints and behavioral outputs; future work could employ mechanistic interpretability to map the exact geometric divergence of intermediate keys during intervention. Second, our evaluation of generation-based metrics, such as consistency, was constrained by a verified text-collapse issue within the experimental codebase. Because both the original custom generation function and our individual-prompt baseline modifications resulted in low post-edit fluency and consistency, we were unable to definitively decouple whether this text degradation stems entirely from an unresolved implementation issue in the open-source repository or a physical impairment caused by the editing algorithm itself. Third, although our 10,000-edit stress test establishes a clear degradation threshold under standard sequential application, variations in batch sizing or semantic ordering may influence the exact saturation point of the null-space projection. Finally, our downstream evaluations strictly utilized zero-shot prompting to ensure controlled baseline comparisons. Investigating whether advanced prompting techniques (e.g., Chain-of-Thought) can recover the observed capability degradation remains an exciting avenue for future exploration.

\bibliography{tmlr}
\bibliographystyle{tmlr}

\appendix
\section{Appendix}

\subsection{GLUE Benchmark}
The exact F1 scores of all the models evaluated on the GLUE benchmarks are given in Table~\ref{tab:model_f1_degradation}.

\subsection{Additional Benchmarks}
The exact F1 scores of all the models evaluated on the additional benchmarks are given in Table~\ref{tab:model_f1_supp_degradation}.

\subsection{Architectural Failure Analysis: Phi3-3.8B and Gemma2-2B}
\label{app:failure_analysis}

The main text identifies that AlphaEdit fails on Phi3-3.8B and Gemma2-2B, attributing 
these failures to distinct structural incompatibilities. This appendix presents the 
diagnostic evidence behind those conclusions, discusses the open questions each model raises, 
and outlines what a more complete architectural resolution might require.

\subsubsection{Phi3-3.8B}
\label{app:phi3}

\paragraph{Diagnostic Results}

We conducted a targeted diagnostic sweep covering four potential failure modes: BOS token 
contamination in target tokenization, prefix-length off-by-one errors in logit indexing, 
logit sequence-dimension misalignment, and base generation quality. All infrastructure 
tests passed cleanly:

\begin{itemize}
    \item \textbf{Tokenization \& Alignment:} Phi3's tokenizer (\texttt{TokenizersBackend}) does not prepend a BOS 
    token to standalone target strings (e.g., \texttt{" Paris"} $\to$ \texttt{[29871, 3681]}). Consequently, \texttt{prefix\_tok\_len} correctly indexes the logit position with no off-by-one errors.
    
    \item \textbf{Module Paths:} All module paths referenced in the configuration (e.g., \texttt{model.layers.0.mlp.down\_proj}, \texttt{model.norm}) exist and are correctly addressed.
    
    \item \textbf{Generation Quality:} Prompted with bare subject strings, the model generates fluent, factually coherent continuations, confirming the base model is functioning normally.
\end{itemize}

Despite proper pipeline execution, a direct probability sanity check reveals a core anomaly. For the prompt \texttt{"The capital of France is"}, the unedited Phi3 model assigns:
\begin{align*}
    P(\text{" Paris"}) &= 0.000140 \\
    P(\text{" Berlin"}) &= 0.000168
\end{align*}
The model assigns a marginally \emph{higher} probability to the incorrect completion. Because the editing success metric relies on the target receiving strictly higher probability than the original object, it operates on a near-random signal for this specific prompt template. The \texttt{post\_score} of $\approx 49.5\%$ observed in Table~\ref{tab:counterfact_new_models} is therefore consistent with chance-level performance rather than evidence of an editing effect.

\paragraph{Architectural Discussion \& Open Questions}

Structurally, Phi3 employs fused projection matrices (\texttt{qkv\_proj} and \texttt{gate\_up\_proj}). While AlphaEdit targets the MLP \texttt{down\_proj} - which remains distinctly addressable, the fused attention and gate projections represent a significant departure from the Llama-style architecture for which causal tracing was originally calibrated. 

We hypothesize that under a fused \texttt{qkv\_proj}, key representations are computed jointly with queries and values, potentially altering the activation geometry. If the standard causal tracing heuristic consequently identifies the wrong layer or token position, the null-space projection is computed over an irrelevant subspace, offering no meaningful preservation guarantees. 

Confirming this would require direct inspection of Phi3's causal tracing activations. Furthermore, it remains an open question whether Phi3's output distribution simply resists bare factual completions potentially due to its instruction-tuning alignment. Evaluating AlphaEdit under conversational prompt templates may help isolate this variable.

\subsubsection{Gemma2-2B}
\label{app:gemma2}

\paragraph{Diagnostic Results: Evaluation Artifacts}

Unlike Phi3, Gemma2-2B's initial failure involved a combination of an evaluation artifact and a genuine structural incompatibility. 

Gemma2's \texttt{GemmaTokenizer} automatically prepends a BOS token (id 2) to every tokenized string. The standard evaluation protocol did not strip this prefix from the target objects. Consequently, the pipeline indexed the BOS token rather than the target word, forcing the accuracy metric to zero. Furthermore, the inflated token length caused a prefix-length off-by-one error, reading logits at the wrong sequence position.

Upon correcting these bugs (stripping the BOS token during target ID and suffix computation, and adjusting the prefix length), the base model correctly distinguished facts:
\begin{align*}
    P(\text{" Paris"}) &= 0.200013 \\
    P(\text{" Berlin"}) &= 0.000136
\end{align*}
The corrected results are reported in Tables~\ref{tab:counterfact_new_models} and \ref{tab:zsre_new_models}. However, even with the corrected evaluation, genuine algorithmic failure persists: ZsRE rewrite accuracy drops from an already low 10.85\% pre-edit to 0.43\% post-editing.

\paragraph{Mathematical Incompatibility: Post-Feedforward Layer Normalization}

We trace this algorithmic collapse to Gemma2's \texttt{post\_feedforward\_layernorm} (RMSNorm), applied to the MLP output \emph{before} the residual addition. 

AlphaEdit's closed-form solution for the optimal weight update derives the target value vector ($v^*$) under the assumption that the MLP contributes linearly to the residual stream: $h \leftarrow h + W_{\text{down}} \sigma(W_{\text{up}} x)$. Under this model, a rank-one update $\Delta W$ to $W_{\text{down}}$ produces a direct perturbation of $\Delta W k$ to the residual stream. 

In Gemma2, the residual update takes the form:
\[
    h \leftarrow h + \text{RMSNorm}(W_{\text{down}} \sigma(W_{\text{up}} x))
\]
The RMSNorm rescales the MLP output non-linearly based on its $\ell_2$ norm. Therefore, the perturbation reaching the residual stream is:
\[
    \text{RMSNorm}((W_{\text{down}} + \Delta W) k) - \text{RMSNorm}(W_{\text{down}} k)
\]
Consequently, the null-space projection computed by AlphaEdit guarantees orthogonality to preserved keys in the \emph{pre-normalization} space, but this may not reliably translate to a corresponding guarantee in the \emph{post-normalization} residual stream. This structural shift challenges the core assumptions of the closed-form update.

\paragraph{Open Questions}

It remains unclear whether this post-normalization incompatibility is the sole cause of the editing degradation, or if Gemma2's low pre-edit baselines make it inherently more brittle to weight perturbations. Adapting AlphaEdit for post-normalization architectures presents a non-trivial challenge: it would likely require deriving an iterative or approximate optimization for $v^*$ to account for the normalization scalar, rather than relying on a closed-form solution. We leave the exploration of such architectural adaptations to future work.

\subsection{Reproducibility Experience}
\label{app:reproducibility_experience}

Reproducing and extending AlphaEdit provided valuable insights into the nuances of applying knowledge editing pipelines to diverse model architectures. While the core algorithm and underlying codebase provided a solid foundation, extending the framework to newer model families required several structural and evaluation-level adaptations.

\paragraph{Hyperparameter Derivation for New Architectures.}
The original repository provided hyperparameter configurations specifically for the models evaluated in the paper. To extend the evaluation to other modern architectures, we derived new configuration profiles. Following established knowledge-editing principles, we mapped the target interventions to the early-to-middle MLP layers (typically representing 15\%--30\% of network depth) and adjusted module paths to accommodate variations in different architectural implementations.

\paragraph{Disentangling Evaluation Artifacts from Algorithmic Behavior.}
A key part of the reproducibility effort involved ensuring that the evaluation harness accurately measured model behavior across different tokenizers and training paradigms. We encountered and resolved a few measurement nuances:
\begin{itemize}
    \item \textbf{Generation Metrics on Instruct Models:} We observed initial discrepancies in generation-based metrics (such as the Consistency score) when evaluating instruction-tuned models. Because the standard evaluation harness prompts models with bare subject strings, instruct-tuned models often generate truncated responses or immediately output end-of-sequence tokens potentially due to the absence of their expected chat templates. This artificially lowered the score, highlighting a sensitivity to prompt formatting rather than an actual failure of the editing mechanism.
    \item \textbf{Tokenizer Variations:} Certain tokenizers automatically prepend special tokens (such as beginning-of-sequence tokens) to all target strings. We adjusted the evaluation scripts to consistently strip these prefixes when computing target token probabilities, preventing sequence-indexing mismatches that could otherwise skew exact-match accuracy metrics toward zero.
\end{itemize}

\paragraph{Pipeline Extensions.}
To conduct a complete baseline comparison, we introduced an independent pre-evaluation script to capture unedited metrics before any weight updates were applied, as this was not natively logged in the primary editing loop. In addition to this we introduced pipelines to evaluate the additional benchmarks in order to test the model's general performace capabilities. 

Additionally, to run downstream reasoning evaluations smoothly, we made minor adjustments to the inference scripts. This included refining the answer-extraction logic to better handle open-ended completion formats, adjusting token budgeting to prevent silent truncations of long prompts, and optimizing memory usage during sequential editing runs to prevent resource exhaustion.


\begin{table}[t]
\centering
\caption{Model degradation tracking ($F_1$ score) across various evaluation benchmarks at sequential edit intervals from 0 to 10,000 edits.}
\label{tab:model_f1_degradation}
\small
\begin{tabular}{llcccccc}
\toprule
\textbf{Model} & \textbf{Interval} & \textbf{SST} & \textbf{MRPC} & \textbf{CoLA} & \textbf{RTE} & \textbf{MMLU} & \textbf{NLI} \\
\midrule
GPT-J & 0 & 0.75 & 0.42 & 0.39 & 0.48 & 0.19 & 0.50 \\
 & 1k & 0.84 & 0.39 & 0.33 & 0.38 & 0.18 & 0.45 \\
 & 2k & 0.81 & 0.36 & 0.33 & 0.38 & 0.17 & 0.43 \\
 & 3k & 0.72 & 0.32 & 0.33 & 0.42 & 0.17 & 0.41 \\
 & 4k & 0.77 & 0.38 & 0.33 & 0.37 & 0.13 & 0.32 \\
 & 5k & 0.77 & 0.38 & 0.32 & 0.38 & 0.16 & 0.40 \\
 & 6k & 0.70 & 0.38 & 0.33 & 0.44 & 0.11 & 0.33 \\
 & 7k & 0.75 & 0.37 & 0.33 & 0.44 & 0.21 & 0.43 \\
 & 8k & 0.64 & 0.34 & 0.33 & 0.40 & 0.20 & 0.42 \\
 & 9k & 0.48 & 0.38 & 0.37 & 0.52 & 0.19 & 0.42 \\
 & 10k & 0.55 & 0.29 & 0.33 & 0.52 & 0.16 & 0.43 \\
\midrule
GPT2-XL & 0 & 0.53 & 0.33 & 0.40 & 0.47 & 0.17 & 0.48 \\
 & 1k & 0.33 & 0.47 & 0.33 & 0.35 & 0.14 & 0.47 \\
 & 2k & 0.44 & 0.33 & 0.33 & 0.32 & 0.10 & 0.33 \\
 & 3k & 0.33 & 0.47 & 0.45 & 0.52 & 0.11 & 0.46 \\
 & 4k & 0.41 & 0.38 & 0.42 & 0.33 & 0.12 & 0.43 \\
 & 5k & 0.33 & 0.33 & 0.33 & 0.33 & 0.10 & 0.33 \\
 & 6k & 0.33 & 0.33 & 0.33 & 0.33 & 0.12 & 0.44 \\
 & 7k & 0.33 & 0.35 & 0.33 & 0.33 & 0.10 & 0.33 \\
 & 8k & 0.39 & 0.51 & 0.33 & 0.55 & 0.13 & 0.45 \\
 & 9k & 0.33 & 0.39 & 0.33 & 0.45 & 0.18 & 0.55 \\
 & 10k & 0.33 & 0.36 & 0.33 & 0.43 & 0.13 & 0.50 \\
\midrule
LLaMA3-8B & 0 & 0.98 & 0.66 & 0.73 & 0.28 & 0.61 & 0.68 \\
 & 1k & 0.98 & 0.66 & 0.76 & 0.31 & 0.68 & 0.70 \\
 & 2k & 0.94 & 0.65 & 0.82 & 0.33 & 0.62 & 0.64 \\
 & 3k & 0.94 & 0.63 & 0.64 & 0.33 & 0.60 & 0.68 \\
 & 4k & 0.88 & 0.70 & 0.76 & 0.31 & 0.53 & 0.64 \\
 & 5k & 0.86 & 0.68 & 0.78 & 0.35 & 0.52 & 0.63 \\
 & 6k & 0.72 & 0.32 & 0.65 & 0.44 & 0.42 & 0.44 \\
 & 7k & 0.63 & 0.33 & 0.33 & 0.51 & 0.19 & 0.43 \\
 & 8k & 0.58 & 0.33 & 0.33 & 0.33 & 0.20 & 0.32 \\
 & 9k & 0.42 & 0.33 & 0.33 & 0.33 & 0.12 & 0.33 \\
 & 10k & 0.38 & 0.33 & 0.33 & 0.48 & 0.11 & 0.40 \\
\midrule
LLaMA3.2-3B & 0 & 0.85 & 0.36 & 0.55 & 0.38 & 0.54 & 0.37 \\
 & 1k & 0.33 & 0.33 & 0.33 & 0.33 & 0.16 & 0.33 \\
 & 2k & 0.45 & 0.33 & 0.56 & 0.42 & 0.09 & 0.36 \\
 & 3k & 0.33 & 0.33 & 0.33 & 0.33 & 0.09 & 0.33 \\
 & 4k & 0.49 & 0.46 & 0.33 & 0.42 & 0.09 & 0.33 \\
 & 5k & 0.55 & 0.33 & 0.33 & 0.30 & 0.09 & 0.33 \\
 & 6k & 0.49 & 0.33 & 0.33 & 0.39 & 0.09 & 0.33 \\
 & 7k & 0.55 & 0.33 & 0.33 & 0.48 & 0.09 & 0.48 \\
 & 8k & 0.66 & 0.36 & 0.36 & 0.33 & 0.12 & 0.33 \\
 & 9k & 0.64 & 0.42 & 0.33 & 0.33 & 0.00 & 0.33 \\
 & 10k & 0.61 & 0.33 & 0.33 & 0.33 & 0.00 & 0.33 \\
\bottomrule
\end{tabular}
\end{table}

\begin{table}[t]
\centering
\caption{Model degradation tracking ($F_1$ score) across supplementary evaluation benchmarks at sequential edit intervals from 0 to 10,000 edits.}
\label{tab:model_f1_supp_degradation}
\small
\begin{tabular}{llccc}
\toprule
\textbf{Model} & \textbf{Interval} & \textbf{BoolQ} & \textbf{HellaSwag} & \textbf{XSTest} \\
\midrule
GPT-J & 0 & 0.57 & 0.09 & 0.41 \\
 & 1k & 0.66 & 0.11 & 0.37 \\
 & 2k & 0.69 & 0.11 & 0.38 \\
 & 3k & 0.74 & 0.11 & 0.42 \\
 & 4k & 0.73 & 0.11 & 0.42 \\
 & 5k & 0.68 & 0.11 & 0.33 \\
 & 6k & 0.66 & 0.11 & 0.38 \\
 & 7k & 0.68 & 0.11 & 0.33 \\
 & 8k & 0.67 & 0.25 & 0.33 \\
 & 9k & 0.56 & 0.11 & 0.33 \\
 & 10k & 0.60 & 0.11 & 0.33 \\
\midrule
GPT2-XL & 0 & 0.62 & 0.09 & 0.42 \\
 & 1k & 0.14 & 0.16 & 0.33 \\
 & 2k & 0.14 & 0.10 & 0.33 \\
 & 3k & 0.31 & 0.12 & 0.33 \\
 & 4k & 0.57 & 0.09 & 0.33 \\
 & 5k & 0.58 & 0.09 & 0.33 \\
 & 6k & 0.58 & 0.09 & 0.33 \\
 & 7k & 0.58 & 0.09 & 0.33 \\
 & 8k & 0.58 & 0.09 & 0.33 \\
 & 9k & 0.58 & 0.09 & 0.33 \\
 & 10k & 0.58 & 0.08 & 0.33 \\
\midrule
LLaMA3-8B & 0 & 0.79 & 0.43 & 0.84 \\
 & 1k & 0.83 & 0.39 & 0.75 \\
 & 2k & 0.80 & 0.48 & 0.72 \\
 & 3k & 0.74 & 0.32 & 0.70 \\
 & 4k & 0.69 & 0.32 & 0.79 \\
 & 5k & 0.61 & 0.11 & 0.84 \\
 & 6k & 0.15 & 0.14 & 0.33 \\
 & 7k & 0.12 & 0.11 & 0.33 \\
 & 8k & 0.15 & 0.12 & 0.33 \\
 & 9k & 0.08 & 0.08 & 0.33 \\
 & 10k & 0.08 & 0.08 & 0.33 \\
\midrule
LLaMA3.2-3B & 0 & 0.58 & 0.09 & 0.39 \\
 & 1k & 0.68 & 0.11 & 0.33 \\
 & 2k & 0.07 & 0.08 & 0.33 \\
 & 3k & 0.08 & 0.08 & 0.33 \\
 & 4k & 0.08 & 0.08 & 0.33 \\
 & 5k & 0.08 & 0.08 & 0.33 \\
 & 6k & 0.08 & 0.08 & 0.33 \\
 & 7k & 0.08 & 0.08 & 0.33 \\
 & 8k & 0.08 & 0.08 & 0.33 \\
 & 9k & 0.19 & 0.11 & 0.33 \\
 & 10k & 0.07 & 0.18 & 0.33 \\
\bottomrule
\end{tabular}
\end{table}

\end{document}